\documentclass[nohyperref]{article}

\usepackage{microtype}
\usepackage{graphicx}
\usepackage{subcaption
}
\usepackage{booktabs} 

\usepackage{hyperref}

 \usepackage[accepted]{icml2022}

\usepackage{amsmath}
\usepackage{amssymb}
\usepackage{mathtools}
\usepackage{amsthm}

\usepackage{multirow}
\usepackage{adjustbox}

\usepackage[capitalize,noabbrev]{cleveref}

\theoremstyle{plain}

\theoremstyle{definition}

\theoremstyle{remark}

\usepackage[textsize=tiny]{todonotes}

\icmltitlerunning{Locally Supervised Learning with Periodic Global Guidance}

\begin{document}

\twocolumn[
\icmltitle{Locally Supervised Learning with Periodic Global Guidance}



\icmlsetsymbol{equal}{*}

\begin{icmlauthorlist}
\icmlauthor{Hasnain Irshad Bhatti}{yyy}
\icmlauthor{Jaekyun Moon}{yyy}
\end{icmlauthorlist}

\icmlaffiliation{yyy}{Department of Electrical Engineering, KAIST, Daejeon, South Korea}

\icmlcorrespondingauthor{Hasnain Irshad Bhatti}{hasnain@kaist.ac.kr}

\icmlkeywords{Machine Learning, ICML}

\vskip 0.3in
]



\printAffiliationsAndNotice{}  

\begin{abstract}

Locally supervised learning aims to train a neural network based on a local estimation of the global loss function at each decoupled module of the network.
Auxiliary networks are typically appended to the modules to approximate the gradient updates based on the greedy local losses.
Despite being advantageous in terms of parallelism and reduced memory consumption, this paradigm of training severely degrades the generalization performance of neural networks.
In this paper, we propose Periodically Guided local Learning (PGL), which reinstates the global objective repetitively into the local-loss based training of neural networks primarily to enhance the model's generalization capability. We show that a simple periodic guidance scheme begets significant performance gains while having a low memory footprint. We conduct extensive experiments on various datasets and networks to demonstrate the effectiveness of PGL, especially in the configuration with numerous decoupled modules. 

\end{abstract}

\section{Introduction}

Backpropagation \cite{backprop, simonyan2014very} has been widely adopted for training Deep Neural Networks (DNNs). Generally, a prediction loss is computed at the end of the network which is backpropagated sequentially through the network while updating the weights of the corresponding layer. Although this update scheme works exceptionally well, it often faces criticism for memory and computation constraints. Firstly, storing the network weights and activations incurs intensive consumption of memory which reduces its potential to be applicable to large networks and on small hardware. Secondly, since the weights of the network should wait for the gradient information before the update, this backward locking makes the training of DNNs more time consuming specially when the depth of the network increases.

Decoupled training of neural networks \cite{Hinton2006AFL,bengioGreedy2006,nokland2019training,belilovsky2019greedy,belilovsky2020decoupled} makes the training procedure more memory efficient as well as allows local parallelization by unlocking the backward pass. Specifically, model is divided into various isolated chunks which are trained independently according to their local loss functions hence can be termed as Locally Supervised Learning (LSL). A layer only has to wait for the activations (or the inputs in the case of first layer) from the previous layer in order to update its weights. Unlike backpropagation (BP), LSL alleviates the need to store the activations for the backward pass hence harness the benefit of having low memory footprint and small computational overhead \cite{wang2021revisiting,laskin2020parallel}. Also, LSL seems to be more biologically plausible than BP \cite{Crick1989,Dan2004,bengioGreedy2006,nokland2019training}. However, LSL is shortsighted (Wang et al., 2021); the update of layers directed by local losses results in significant information loss at every consecutive chunk of the network. This shortsightedness of the network limits the information propagation within the network resulting in poor prediction performance as compared to BP. 

Recent works incorporated additional objective functions to account for the lost information in LSL. \cite{nokland2019training} utilizes a similarity measure in addition to simple prediction loss, hence leading each layer towards an objective which is more globally aware. However, the generalization performance for large networks and big datasets still suffers. \cite{wang2021revisiting} used the mutual information based objective function to retain the task-relevant information throughout the network. Although the technique show competitive performance, it induces computation complexity and large memory footprint because of additional large decoder networks. 

\begin{figure*}[t]

\begin{center}
\vskip 0.2in
\begin{subfigure}[b]{0.4\textwidth}
           \centering
         \includegraphics[width=\textwidth]{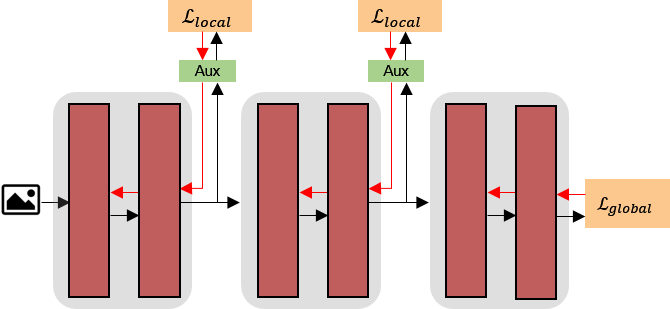}
	\caption*{Greedy Local Update}
     \end{subfigure}
     \hspace{1cm} 
  \begin{subfigure}[b]{0.4\textwidth}
           \centering
	     	 
      	 \includegraphics[width=\textwidth]{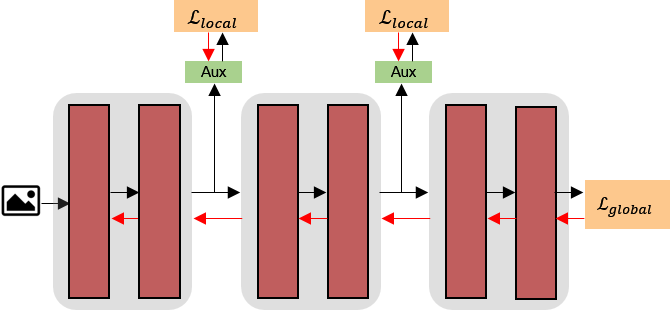}
	\caption*{Periodic Global Guidance, $Q$ epochs if $epoch \% P=0$}
     \end{subfigure}

\caption{Proposed Methodology: Left figure shows the local-loss based update, where a greedy estimate of loss function guides the module update.
Right figure shows PGL. During the periodic global guidance period P, the network is updated based on two loss functions. 1) the local loss function updates only the auxiliary network, and 2) the global loss function computed at the end of the network backpropagates through the whole network to perform a global-guided update.} \label{proposedmetho}
\end{center}
\vskip -0.2in
\end{figure*}

\textbf{Contributions:} In this paper, we propose a new strategy for training decoupled neural networks, PGL, which is based on the periodic guidance of decoupled modules with the global objective. Specifically, we introduce an alignment period in which the global loss of the network provides a gradient update to the network, whereas the network is updated with local losses otherwise. Empirically, we show that this periodic guidance of decoupled modules results in better alignment of the greedy objectives with the global objective, resulting in a better generalization performance. Extensive experimentation with various networks and datasets show that our method achieves better performance than the baseline decoupled training. Detailed studies on choosing the size of auxiliary networks and tuning the global guidance period show that we can further reduce the performance gap between the decoupled training of neural networks and the backpropagation.

\section{Proposed Methodology}

We train a neural network in a block-wise fashion similar to DGL \cite{belilovsky2020decoupled}. Auxiliary networks are attached to each block of the network to estimate a local loss. However, instead of just relying on the greedy objectives of the network, we introduce a notion of periodic global guidance. The motivation is to instill global information to each block of the network periodically with period $P$, such that the greedy objectives at each block of the network get aligned to a more globally aware objective. Overall, we only perform this global guidance step for a short duration $Q$ (1 or 2 epochs) after the local-loss based update. We assert that, by performing this periodic global guidance, PGL increases the generalization performance of decoupled training significantly without adding large decoder networks \cite{wang2021revisiting} or ineffective objectives \cite{nokland2019training}. We define our problem statement as follows:

Consider a neural network containing a total of $K$ layers is divided in $J$ number of blocks, where each $j^{th}$ block $(1\le j\le J)$ is composed of ${\frac{K}{J}}$ layers and ${K\ge J}$. Let the inputs to the network be ${X}_0$ and corresponding labels as $Y$. The output of each block ${X}_j$ is given by ${X}_{j} =f_{\theta_{j}}({X}_{j-1})$ where $f_{\theta_{j}}$ represents the function of a block $j$ having weight parameters $\theta_{j}$. Consider an auxiliary layer with parameters $\gamma_{j}$ attached to each block $j$ which outputs the prediction logits $p_j$ at each layer according to $p_j = f_{\gamma_{j}}({X}_{j})$. The proposed training procedure of PGL is depicted in Figure \ref{proposedmetho}.

\subsection{Layer-wise Prediction Loss}
For each block $j$ of the network, a local greedy loss $\mathcal{L}_{local}$ is calculated based on the true labels of the input data. The inputs ${X}_{j-1}$ are processed through block $j$ with parameters $\theta_{j}$ to get representations ${X}_{j}$ and the auxiliary network $f_{\gamma_{j}}$ to obtain auxiliary predictions. These predictions are utilized to compute the local loss i.e. cross-entropy loss as follows:
\begin{equation}
\label{eqn:localloss}
\mathcal{L}_{local} = CrossEntropy(Y, f_{\gamma_{j}}(f_{\theta_{j}}({X}_{j-1}))
\end{equation} 
The cross-entropy loss for each block is the same as in ~\cite{belilovsky2020decoupled,nokland2019training,wang2021revisiting}. The blocks of the networks are updated based on the following optimization: 
\begin{equation}
\label{eqn:locallossopt}
\min_{\theta_j,\gamma_{j}}\mathcal{L}_{local}(Y, {X}_{j-1}; \gamma_{j}, \theta_{j})
\end{equation}
Minimizing this loss means that each block of the network learns a greedy mapping from its input space to the label space according to the capacity of the block alongwith its auxiliary network. Consequently, this loss will be higher for initial blocks of the network because of its small size. In spite of that, greedy objectives at initial blocks of the network contributes little to improving the hierarchical nature of representations and thus degrades the overall generalization performance of the network.

\subsection{Periodic Guidance with Global Error Signals}
Decoupled learning originally is proposed to reap benefits of low memory consumption, pipelining and parallelization as compared to back-propagation. However, training neural networks based on local losses suffers from severe performance degradation. Recent works reduce the performance gap to an extent but at a cost of enormous additional computation. This motivates us to look for other methods to enhance generalization performance. Here, we present a novel technique of periodically guiding the greedy objectives to a global objective which increases the generalization performance of the network significantly while having a low-memory footprint.

\begin{figure}[t]

\begin{center}
\begin{subfigure}[b]{0.4\textwidth}
	\includegraphics[width=\textwidth]{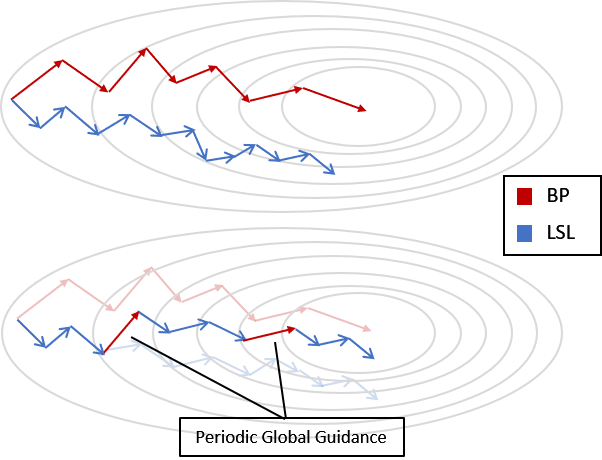}
     \end{subfigure}

\caption{This illustration depicts the effect of periodic global guidance, which leads to a solution closer to that of global-loss-based Backpropagation.} \label{loss}
\end{center}
\vskip -0.2in
\end{figure}

We propose to allow the global loss calculated at the end of the network to periodically guide the network parameters to a better update direction. This guidance step is performed at a certain period $P$. Figure \ref{loss} illustrates the effect of periodic global guidance on the loss landscape.

The network first starts training layer-wise by using the greedy cross-entropy loss generated at each auxiliary layer as defined in (\ref{eqn:localloss}). And updating the network using local losses for $P$ epochs, the whole network is updated based on the global loss function as follows:
\begin{equation}
\label{eqn:globalloss}
\min_{\theta_1, ... , \theta_J}\mathcal{L}_{global}(Y, f_{\theta_{j}}(f_{\theta_{j-1}}(,...,(f_{\theta_{1}}({X}_{0})))))
\end{equation} 
The parameters $\gamma_{j}$ are not updated as the auxiliary networks lie outside the global backpropagation path. Instead, we update the auxiliary networks using local losses as follows:
\begin{equation}
\label{eqn:auxnetopt}
\min_{\gamma_{j}}\mathcal{L}_{local}(Y, {X}_{j-1}; \gamma_{j}, \theta_{j})
\end{equation}

\vskip -0.2in

\subsection{Proxy Gradient Estimator and \textit{AUX-ADAPT}}
Auxiliary network attached to a module can be regarded as a proxy gradient estimator to the rest of the network i.e. for a module $i$, $f_{\gamma_{i}}$ acts as a proxy for $\forall f_{\theta_{i < j \leq J}} $ since it provides an update based on the same loss function as the rest of the network, except $\gamma_{j}$ is smaller in size. Intuitively, $f_{\gamma_{j}}$ with the initial modules of the network provide poor proxy than the later ones. Thereby, we propose to use adaptive auxiliary network \textit{\small{AUX-ADAPT}}, where large auxiliary network is attached with initial modules of the network and gradually its size decreases as we reach to the final module.

\section{Experiments and Results}\label{chapterEnR}

\subsection{Experimental Setup}

We perform experiments using ResNet-32 and ResNet-110 \cite{He_2016_CVPR} on a variety of datasets, such as CIFAR-10 \cite{krizhevsky2009learning}, SVHN \cite{netzer2011reading} and STL-10 \cite{pmlr-v15-coates11a}. SGD optimizer is used with a Nesterov momentum of 0.9 and L2 weight decay ratio of 1e-4. We used the batch size of 1024 for CIFAR-10 and SVHN, and 128 for STL-10 and trained the network for 160 epochs. Initial learning rate is set to 0.8 and it follows cosine annealing. We set the period of global guidance, $P=10$ and perform $Q = 2$ epochs of global loss guided update. 

\begin{table*}[!t]
\vskip 0.1in
		\caption{\small Generalization performance of PGL with difference number of decoupled blocks ($J = 16$, $J = 8$, $J = 4$, $J = 2$) using ResNet architecture on CIFAR-10, SVHN and STL-10. The avarage test accuracy is reported after 3 independent trials. In every setting, PGL achieves significantly higher accuracy as compared to DGL specially with large number of decoupled blocks in the network. }
\vskip 0.1in
	\centering
		
	\label{table:resultswithblocks}
	\begin{sc}
	\begin{adjustbox}{width=1.5\columnwidth,center}
	\begin{tabular}{lll||cccc|c}
		\toprule  
 
		& & & \multicolumn{4}{c}{\textbf{Network Blocks}} & \\
	\cmidrule{4-8}
	{\textbf{Dataset}} & {\textbf{Network}}	& {\textbf{Methods}} & $J = 16$ &$J = 8$	& $J = 4$ & $J = 2$ & \textbf{Backprop}\\
		
		\midrule
		\multirow{9}{*}{CIFAR-10} & \multirow{5}{*}{ResNet32} 
			& DGL (baseline) 	&	83.63\% 		&	85.70\% 		&	88.52\%		&	91.29\%		& \multirow{5}{*}{93.03\%}	\\	
		 &	& InfoPro 			&	85.04\%		&	88.94\%		&	91.12\%		&	91.81\%		& 							\\
		 &	& PGL (1Conv-2FC) 	& 	87.26\%		&	88.30\%		&	89.72\%		&	91.89\%		&							\\
		 &	& PGL (aux-adapt) 	& 	88.85\%		&	89.81\%		&	90.86\%		&	92.01\%		&							\\
		 &	& PGL (2Conv-2FC) 	& 	90.22\%		&	90.55\%		&	91.62\%		&	92.59\%		&							\\
		
		\cmidrule{2-8}
		& \multirow{4}{*}{ResNet110} 
			& DGL (baseline) 	& 	86.12\% 		&	87.55\% 		&	89.59\%		&	92.38\%		& \multirow{4}{*}{93.39\%}	\\
		 &	& InfoPro 			& 	88.75\% 		&	90.75\% 		&	91.06\%		&	92.38\%		& 							\\
		 &	& PGL (1Conv-2FC) 	& 	88.29\% 		&	89.86\% 		&	91.36\%		&	92.88\%		& 							\\
		 &	& PGL (aux-adapt) 	& 	90.37\% 		&	90.94\% 		&	91.86\%		&	93.20\%		& 							\\
		
		\midrule
		\multirow{3}{*}{SVHN}& \multirow{3}{*}{ResNet110} 
			&	DGL (baseline) 	& 	94.64\% 		& 	94.61\%		&	95.53\%		&	96.40\% 		& \multirow{3}{*}{96.96\%}	\\	
		 &	& 	InfoPro 			& 	94.84\% 		&	95.97\% 		&	96.30\%		&	96.72\%		& 							\\
		 & 	&	PGL (1Conv-2FC)	& 	95.37\% 		& 	95.33\%		&	95.63\% 		&	96.69\% 		&							\\
		 & 	&	PGL (aux-adapt) 	& 	95.62\% 		& 	95.71\%		&	96.02\% 		&	96.77\% 		& 							\\
		
		\midrule		
		\multirow{3}{*}{STL-10}& \multirow{3}{*}{ResNet110} 
			&	DGL (baseline) 	& 	72.46\% 		&	73.42\% 		&	72.65\% 		&	75.80\%		& \multirow{3}{*}{78.13\%}	\\	
		 & 	&	PGL (1Conv-2FC)	&	73.76\% 		&	74.96\% 		&	74.92\%  	&	76.02\%		&							\\ 			
		 & 	&	PGL (aux-adapt) 	&	75.19\% 		&	75.62\% 		&	76.17\%  	&	76.14\%		&		 					\\ 			
		\bottomrule
	\end{tabular}
	\end{adjustbox}
	\end{sc}

	\vspace{-1mm}
\end{table*}

\vspace{4pt}
\textbf{Auxiliary Layer Setting:}
We follow the setting of the splitting our model according to \cite{wang2021revisiting}. We consider splitting a network into 16, 8, 4 and 2 modules. We attach different sizes of auxiliary networks; \textit{1Conv-1FC}, \textit{2Conv-2FC} setting contains 1 and 2 convolutional layers followed by 1 and 2 fully connected layers respectively, while our proposed \textit{\small{AUX-ADAPT}} contains \textit{2Conv-2FC, 1Conv-3FC and 1Conv-2FC} networks for input channels 16, 32 and 64 respectively.

\begin{table*}[!t]
\vskip 0.1in
		\caption{\small Performance against network size and memory footprint. Total parameters of ResNet-32 including various auxiliary networks is reported. MEM represents the GPU memory utilization during training. Nvidia GeForce RTX 2080 Ti GPU was used for all experiments.}	
\vskip 0.1in
	\centering
	\label{table:resultswithauxsize}
	
	\begin{sc}
	\begin{adjustbox}{width=2.1\columnwidth,center}
	\begin{tabular}{lcccccccccccc}
		\toprule  

 		 \multicolumn{1}{c}{\multirow{3}{*}{\textbf{Method}}} & \multicolumn{12}{c}{\textbf{Network Blocks}}  \\

		& \multicolumn{3}{c}{J = 16}	& \multicolumn{3}{c}{J = 8} & \multicolumn{3}{c}{J = 4} & \multicolumn{3}{c}{J = 2}\\ 	   									
		
		\cmidrule(lr){2-4}  \cmidrule(lr){5-7} \cmidrule(lr){8-10} \cmidrule(lr){11-13}		
		&Test Acc & 	Param & Mem	& Test Acc 	 & 	Param& Mem &  Test Acc &	Param& Mem  &	Test Acc & Param& Mem  \\	
		
		\midrule
		 DGL (baseline)		& 83.62\% & 	0.86M & 1.62 GB 	&85.70\% 	& 0.65M&	 1.87 GB 	&	88.52\%	& 0.55M&	 2.18 GB &	91.29\%	& 0.50M& 3.02 GB 	\\
		 InfoPro (SOTA)		& 85.04\% & 	0.92M &	2.56 GB &88.94\%		& 0.68M& 2.96 GB 	&	91.1\%	& 0.57M&	 3.18 GB &	91.81\%	& 0.50M& 3.29 GB 	\\
		 PGL (1Conv-2FC)  	& 87.26\% & 	0.86M &	2.03 GB &88.29\%  	& 0.65M& 2.23 GB 	& 	89.72\%	& 0.55M& 2.48 GB &	91.89\%	& 0.50M&	 3.15 GB		 \\
 		 PGL (aux-adapt)		& 88.85\% & 	0.98M &	2.08 GB &89.81\%  	& 0.71M& 2.30 GB 	& 	90.86\%	& 0.58M& 2.55 GB &	92.01\%	& 0.51M& 3.17 GB 	\\
		 PGL (2Conv-2FC)		& 90.22\% & 	1.24M &	2.12 GB &90.55\%  	& 0.82M& 2.31 GB 	& 	91.61\%	& 0.64M& 2.56 GB &	92.59\%	& 0.53M& 3.17 GB 	\\
				\midrule
		 Backprop			& 93.4\% & 	0.86M &	3.57 GB	&-&-&-&-&-&-&-&-&-	\\
		\bottomrule
	\end{tabular}
	\end{adjustbox}
	\end{sc}
	\vspace{-1mm}
\end{table*}

\subsection{Results and Discussion}
We compare our approach to the baseline DGL \cite{belilovsky2020decoupled} and SOTA InfoPro \cite{wang2021revisiting} in terms of generalization performance and memory advantage. As reported in Table \ref{table:resultswithblocks}, the proposed PGL outperforms baseline DGL in all the training scenarios and results in test accuracy higher or almost on par as compared to the InfoPro with our proposed \textit{\small{AUX-ADAPT}} setting. In the case of CIFAR-10 trained with ResNet-32, the generalization gap of PGL with $J=16$ blocks is remarkably 3.63\% and 6.59\% against DGL while 2.22\% and 3.81\% as compared to InfoPro with auxiliary networks \textit{\small{1Conv-2FC}} and \textit{\small{AUX-ADAPT}} respectively. Using 2 Conv layers in the auxiliary network, PGL achieves highest accuracy of 92.59\% and 93.20\%, only 0.44\% and 0.19\% less than that of BP with ResNet-32 and ResNet-110 respectively. Similarly for SVHN and STL-10, PGL generalizes better than DGL, resulting in significant performance gain specially with large $K$.

To understand the hardware complexity, we report the network parameters and memory footprint (average GPU memory usage) of PGL in Table \ref{table:resultswithauxsize}. It can be observed that the average memory usage of PGL lies in between DGL and InfoPro. For instance, PGL with \textit{\small{AUX-ADAPT}} enjoys significant accuracy gain of 5.22\%, 4.11\%, 2.34\% and 0.72\% as compared to DGL while using 42\%, 35\%, 28\% and 11\% less memory as compared to BP for 16, 8, 4 and 2 blocks of the network respectively. Comparatively, InfoPro consumes significantly higher memory than PGL in all cases, due to its expensive computation in the estimation of mutual information. It can be seen that PGL with large auxiliary network still has less memory footprint as compared to BP despite having more parameters. This is because only a single decoupled module alongwith its auxiliary network is loaded into memory at a time as opposed to BP which holds intermediate activations of the whole network in memory. 

We provide an ablation study of global guidance period in Table \ref{table:effectofperiod}. ResNet-32 with $J=16$ with \textit{\small{AUX-ADAPT}} is trained for various periods i.e. $P = 5, 10, 15$ and $20$ with durations $Q = 1, 2$ and $3$. Accuracy gain can be noted in all settings; about $7.3\%$ with the most periodic setting and $Q=3$ while $3.7\%$ with the least one $P=20$ and $Q=1$. 

\begin{table}[!t]
	\caption[Change in Alignment Period]{\small Effect of changing guidance period  }
	\centering
	\label{table:effectofperiod}
	
	\begin{sc}
	\begin{adjustbox}{width=0.8\columnwidth,center}	
	
	\begin{tabular}{c||cccc}
		\toprule  
 
		  & \multicolumn{4}{c}{\textbf{Network Blocks}}  \\
		\cmidrule{2-5}
	  PGL (aux-adapt) & $P = 5$ &$P = 10$	& $P = 15$ & $P = 20$ \\
		
		\midrule
			$Q = 1$ & 88.57\% & 87.94\% &	87.84\% & 87.33\% \\			 	
		 	$Q = 2$ & 90.43\% & 88.85\% &	88.37\% & 88.08\%   \\
		 	$Q = 3$ & 90.93\% & 89.52\% &	89.31\% & 88.57\%   \\
		 \midrule
		 	DGL  &  \multicolumn{4}{c}{83.63\%}   \\

		\bottomrule
	\end{tabular}
	\end{adjustbox}
	\end{sc}
	\vspace{-2mm}
\end{table}

\section{Conclusion and Future Work}
We proposed a novel solution PGL to the performance degradation issue in the locally-supervised training of neural networks by introducing a periodic guidance step. Based on our approach, greedy objectives of decoupled modules are aligned to the global objective of the network after a certain period of local updates. Extensive experimentation show that PGL generalizes better than the state-of-the-art techniques while having a lower memory footprint. 

Future works include implementation of PGL with layer-wise pipelining and data parallelism to reduce the latency among codependent modules. Furthermore, auxiliary networks will be designed to have low parameter count e.g. sparse networks to accelerate training. Also, we plan to quantize the periodic global updates to reduce memory footprint and speed-up training.

\section{Acknowledgments}
This work was supported by Institute of Information \& Communications Technology Planning \& Evaluation(IITP) grant funded by the Korea government(MSIT) (No. 2020-0-00626, Ensuring high AI learning performance with only a small amount of training data) and National Research Foundation of Korea(NRF) grant funded by the Korea government(Ministry of Science, ICT \& Future Planning) (No. 2019R1I1A2A02061135).

\bibliography{pgl}
\bibliographystyle{icml2022}

\end{document}